\documentclass[english]{article}
\usepackage{tipa}
\usepackage[utf8]{inputenc}
\usepackage[T1]{fontenc}
\usepackage{babel}
\usepackage{amsmath}
\usepackage{amssymb}
\usepackage{graphicx}
\usepackage{fancyhdr}
\usepackage{booktabs}
\usepackage{underscore}
\usepackage{colortbl,xcolor}
\usepackage{url}

\definecolor{ourred}{HTML}{E18F7E}

\begin{document}
\title{Global-scale phylogenetic linguistic inference from lexical resources}

\author{Gerhard J\"ager\textsuperscript{1}}

\maketitle
\thispagestyle{fancy}

1. T\"ubingen University, Institute of Linguistics, Wilhelmstr.\ 19, 72074 T\"ubingen,
Germany (gerhard.jaeger@uni-tuebingen.de)

\begin{abstract}
  Automatic phylogenetic inference plays an increasingly important role in computational
  historical linguistics. Most pertinent work is currently based on \emph{expert cognate
    judgments}. This limits the scope of this approach to a small number of well-studied
  language families.

  We used machine learning techniques to compile data suitable for phylogenetic inference
  from the ASJP database, a collection of almost 7,000 phonetically transcribed word lists
  over 40 concepts, covering two third of the extant world-wide linguistic diversity.

  First, we estimated \emph{Pointwise Mutual Information} scores between sound classes
  using weighted sequence alignment and general-purpose optimization. From this we
  computed a dissimilarity matrix over all ASJP word lists. This matrix is suitable for
  \emph{distance-based} phylogenetic inference.

  Second, we applied \emph{cognate clustering} to the ASJP data, using supervised training
  of an SVM classifier on expert cognacy judgments.

  Third, we defined two types of binary \emph{characters}, based on automatically inferred
  cognate classes and on sound-class occurrences.

  Several tests are reported demonstrating the suitability of these characters for
  \emph{character-based} phylogenetic inference.
\end{abstract}

\section*{Background \& Summary}

The cultural transmission of natural languages with its patterns of near-faithful
replication from generation to generation, and the diversification resulting from
population splits, are known to display striking similarities to biological evolution
\cite{atkinsonGray05,levinsonGray12}. The mathematical tools to recover evolutionary
history developed in computational biology --- phylogenetic inference --- play an
increasingly important role in the study of the diversity and history of human
languages. \cite{grayJordan2000,dunn2005,pageletal2007,brownetal08,grayDrummondGreenhill09,
  dunnetal11,bouckaertetal12,bowernAtkinson2012,bouchardetal13,pagel2013,hruschkaetal15,jaeger15pnas}

The main bottleneck for this research program is the so far still limited availability of
suitable data. Most extant studies rely on manually curated collections of expert
judgments pertaining to the cognacy of core vocabulary items or the grammatical
classification of languages. Collecting such data is highly labor intensive. Therefore
sizeable collections currently exist only for a relatively small number of well-studied
language
families. \cite{dunnetal11,bouchardetal13,abvd,wichmannHolman13,list14Data,mennecieretal16}

Basing phylogenetic inference on expert judgments, especially judgments regarding the
cognacy between words, also raises methodological concerns. The experts making those
judgments are necessarily historical linguists with some prior information about the
genetic relationships between the languages involved. In fact, it is virtually impossible
to pass a judgment about cognacy without forming a hypothesis about such relations. In
this way, data are enriched with prior assumptions of human experts in a way that is hard
to control or to precisely replicate.

Modern machine learning techniques provide a way to greatly expand the empirical base of
phylogenetic linguistics while avoiding the above-mentioned methodological problem. 

The \emph{Automated Similarity Judgment Program} (ASJP) \cite{asjp17} database contains
40-item core vocabulary lists from more than 7,000 languages and dialects across the
globe, covering about 75\% of the extant linguistic diversity. All data are in phonetic
transcription with little additional annotations.\footnote{The only expert judgments
  contained in the ASJP data are rather unsystematic manual identifications of loan
  words. This information is ignored in the present study.} It is, at the current time,
the most comprehensive collection of word lists available.

Phylogenetic inference techniques comes in two flavors, \emph{distance-based} and
\emph{character-based} methods. Distance-based methods require as input a matrix of
pairwise distances between taxa. Character-based methods operate on a character matrix,
i.e.\ a classification of the taxa under consideration according to a list of discrete,
finite-valued characters. While some distance-based methods are computationally highly
efficient, character-based methods usually provide more precise results and afford more
fine-grained analyses.

The literature contains proposals to extract both pairwise distance matrices and character
data from phonetically transcribed word
lists. \cite{jaeger13ldc,jaegerSofroniev16Konvens,jaegerListSofroniev17} In this paper we
apply those methods to the ASJP data and make both a distance matrix and a character
matrix for 6,892 languages and dialects\footnote{These are all languages in ASJP v.\ 17
  except reconstructed, artificial, pidgin and creole languages.} derived this way
available to the community. Also, we demonstrate the suitability of the results for
phylogenetic inference.

While both the raw data and the algorithmic methods used in this study are freely publicly
available, the computational effort required was considerable (about ten days computing
time on a 160-cores parallel server). Therefore the resulting resource is worth publishing
in its own right.

\section*{Methods}

\subsection*{Creating a distance matrix from word lists}

In \cite{jaeger13ldc} a method is developed to estimate the dissimilarity between two ASJP
word lists. The main steps will be briefly recapitulated here.

\subsubsection*{Pointwise Mutual Information}

ASJP entries are transcribed in a simple phonetic alphabet consisting of 41 sound classes
and diacritics. In all steps described in this paper, diacritics are removed.\footnote{For
  instance, a sequence \texttt{th\textasciitilde}, indicating an aspirated ``t'', is
  replaced by a simple \texttt{t}.} This way, each word is represented as a sequence over
the 41 ASJP sound classes.

The \emph{pointwise mutual information} (PMI) between two sound classes is central for
most methods used in this paper. It is defined as
\begin{equation}
  \label{eq:pmi}
 \mathit{PMI}(a,b) \doteq -\log\frac{s(a,b)}{q(a)q(b)},
\end{equation}
where $s(a,b)$ is the probability of an occurrence of $a$ to participate in a regular
sound correspondence with $b$ in a pair of cognate words, and $q(x)$ are the probabilities
of occurrence of $x$ in an arbitrarily chosen word.

Let ``\texttt{-}'' be the gap symbol. A pairwise alignment between two strings $(x,y)$ is
a pair of strings $(x',y')$ over sound class symbols and gaps of equal length such that
$x$ is the result of removing all gap occurrences in $x'$, and likewise for $y'$. A licit
alignment is one where a gap in one string is never followed by a gap in the other
string. There are two parameters $gp_1$ and $gp_2$, the \emph{gap penalties} for opening
and extending a gap. The aggregate PMI of an alignment is
\begin{equation}
  \mathit{PMI}(x',y') = \sum_i \mathit{PMI}(x'_i,y'_i),
\end{equation}
where $\mathit{PMI}(x'_i,y'_i)$ is the corresponding gap penalty if $x'_i$ or $y'_i$ is a gap.

For a given pair of ungapped strings $(x,y)$, $\mathit{PMI}(x,y)$ is the maximal aggregate
PMI of all possible licit alignments between $x$ and $y$. It can efficiently be computed
with a version of the Needleman-Wunsch algorithm \cite{needlemanWunsch}. In this study, we
used the function \texttt{pairwise2.align.globalds} of the \emph{Biopython} library
\cite{biopython} for performing alignments and computing PMI scores between strings.

\subsection*{Parameter estimation}

The probabilities of occurrence $q(a)$ for sound classes $a$ are estimated as relative
frequencies of occurrence within the ASJP entries. The scores $\mathit{PMI}(a,b)$ for
pairs of sound classes $(a,b)$ and the gap penalties are estimated via an iterative
procedure. 

In a first step, pairwise distances between languages are computed via the method
described in the next subsection, using $1-\mathit{LDN}(x,y)$ instead of
$\mathit{PMI}(x,y)$ as measure of string similarity, where $\mathit{LDN}(x,y)$ is the
\emph{normalized Levenshtein distance} \cite{holmanetal08} between $x$ and $y$, i.e.\ the
edit distance between $x$ and $y$ divided by the length of the longest string. All pairs
of languages\footnote{For the sake of readability, I will use the term ``language'' to
  refer to languages proper and to dialects alike; ``doculect'' would be a more correct if
  cumbersome term.} $(l_1,l_2)$ with a distance $\leq 0.7$ are considered as
\emph{probably related}. This is a highly conservative estimate; $99.9\%$ of all probably
related languages belong to the same language family and about $60\%$ to the same
sub-family.

Next, for each pair of probably related languages $(l_1,l_2)$ and each concept $c$, each
word for $c$ from $l_1$ is aligned to each word for $c$ from $l_2$. The pair of words with
the lowest $\mathit{LDN}$ score is considered as \emph{potentially cognate}.

All pairs of potentially cognate words are aligned using the Levenshtein algorithm, and
for each pair of sound classes $(a,b)$, $s_0(a,b)$ is estimated as the relative frequency
of $a$ being aligned to $b$ across all such alignments. Alignments to gaps are excluded
from this computation. $\mathit{PMI}_0(a,b)$ is then calculated according to
(\ref{eq:pmi}).

Suppose gap penalties $gp_1,gp_2$ and a threshold parameter $\theta$ are given. The final
PMI scores are estimated using an iterative procedure inspired by the \emph{Expectation
  Maximization} algorithm \cite{em}:

\begin{itemize}
\item For $i$ in $1\ldots 10$:
  \begin{enumerate}
  \item All potential cognate pairs are aligned using the $\mathit{PMI}_{i-1}$-scores.
  \item $s_i(a,b)$ is estimated as the relative frequency of $a$ aligned with $b$ among
    all alignments between potential cognates $x,y$ with $\mathit{PMI}_{i-1}(x,y)$ $\geq$
    $\theta$.
  \item $\mathit{PMI}_i$ is calculated using formula (\ref{eq:pmi}).
  \end{enumerate}
\end{itemize}

The \emph{target function} $f(gp_1,gp_2,\theta)$ is the average distance between all
probably related languages using the $\mathit{PMI}_{10}$-scores. The values for
$gp_1,gp_2,\theta$ are determined as those minimizing $f$, using Nelder-Mead optimization
\cite{neldermead}. The following optimal values were found: $gp_1\approx -2.330,
gp_2\approx-1.276, \theta\approx 4.401$.

The threshold $\theta\approx 4.401$ ensures that only highly similar word pairs are used
for estimating PMI scores. For instance, between French and Italian only five word pairs
have a PMI similarity $\geq \theta$ according to the final scores: \emph{soleil}
\texttt{[sole]} -- \emph{sole} \texttt{[sole]} (`sun'; PMI=11.6), \emph{corne}
\texttt{[korn]} -- \emph{corno} \texttt{[korno]} (`horn'; PMI=7.7), \emph{arbre}
\texttt{[arbr3]} -- \emph{albero} \texttt{[albero]} (`tree'; PMI=7.1), \emph{nouveau}
\texttt{[nuvo]} -- \emph{nuovo} \texttt{[nwovo]} (`new'; PMI=7.0), and \emph{montagne}
\texttt{[motaj]} -- \emph{montagna} \texttt{[monta5a]} (`mountain'; PMI=4.9).

\begin{figure}[h!]
  \centering
  \includegraphics[width=\linewidth]{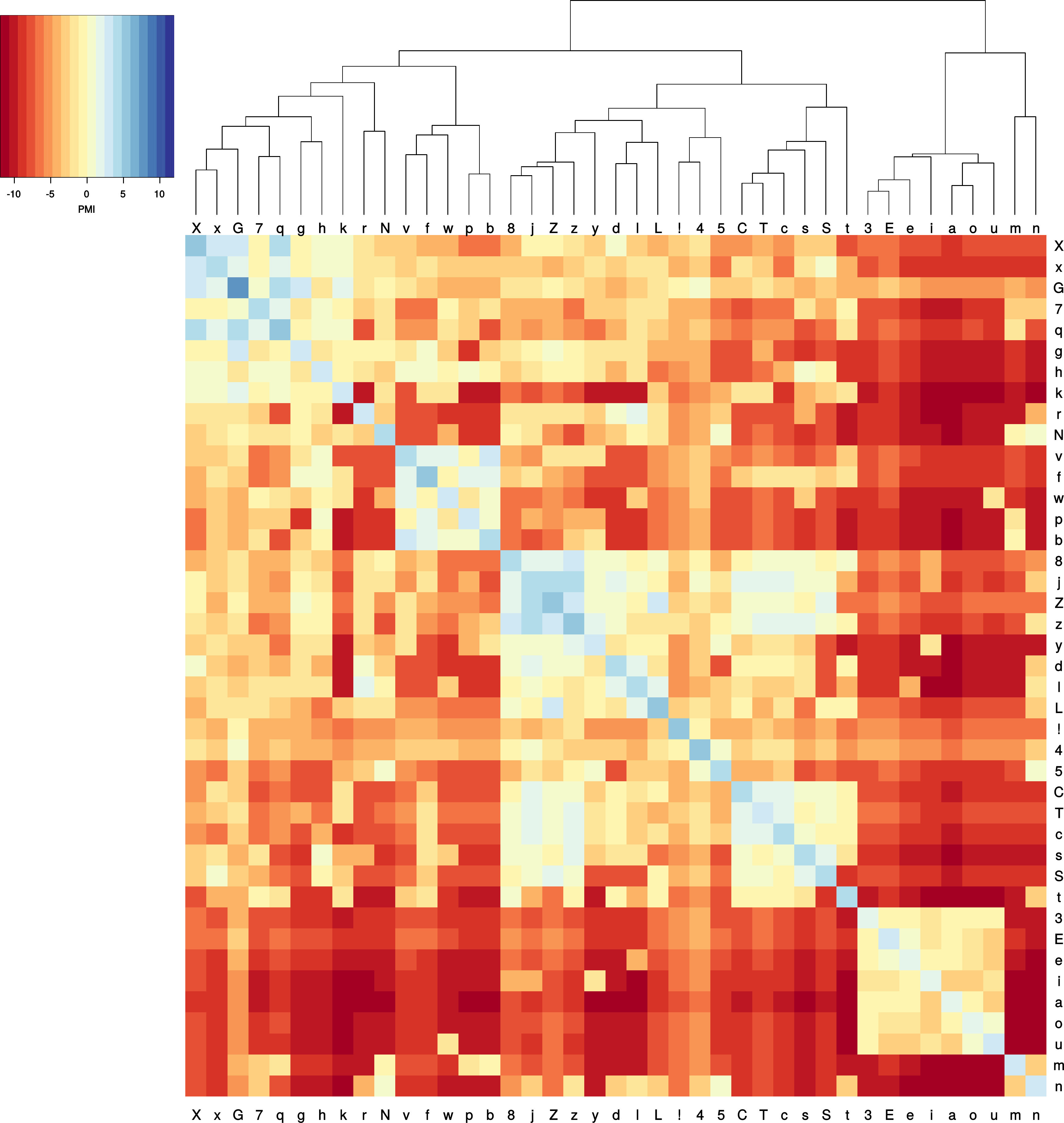}
  \caption{PMI scores. Heatmap and hierarchical clustering dendrogram}
  \label{fig:1}
\end{figure}

The final PMI scores between sound classes are visualized in Figure \ref{fig:1}.
It is easy to discern that $\mathit{PMI}(a,a)$ is positive for all sound classes $a$, and
that $\mathit{PMI}(a,b)$ for $a\neq b$ is negative in most cases. There are a few pairs
$a,b$ with positive score, such as \texttt{b}/\texttt{f}. Generally, sound class pairs
with a similar place of articulation tend to have relatively high scores. This pattern is
also visible in the cluster dendrogram. We observe a primary split between vowels and
consonants. Consonants are further divided into labials, dentals, and velar/uvular sounds.

\subsection*{Pairwise distances between languages}

When aggregating PMI similarities between individual words into a distance measure between
word lists, various complicating factors have to be taken into consideration:
\begin{itemize}
\item Entries for a certain language and a certain concept often contain several
  synonyms. This is a potential source of bias when averaging PMI similarities of
  individual word pairs.
\item Cognate words tend to be more similar than non-cognate ones. However, the average
  similarity level between non-cognate words depends on the overall similarity between the
  sound inventories and phonotactic structure of the languages compared. To assess the
  informativeness of a certain PMI similarity score, it has to be calibrated against the
  overall distribution of PMI similarities between non-cognate words from the languages in
  question.
\item Many ASJP word lists are incomplete, so the word lists are of unequal length.
\end{itemize}

To address the first problem, \cite{jaeger13ldc} defined the similarity score between
languages $l_1$ and $l_2$ for concept $c$ as the maximal PMI similarity between any pair
of entries for $c$ from $l_1$ and $l_2$.

The second problem is addressed by estimating, for each concept $c$ for which both
languages have an entry, the $p$-value for the null hypothesis that none of the words for
$c$ being compared are cognate. This is done in a parameter-free way. For each pair of
concepts $(c_1,c_2)$, the PMI similarities between the words for $c_1$ from $l_1$ and the
words for $c_2$ from $l_2$ are computed. The maximum of these values is the similarity
score for $(c_1,c_2)$. Under the simplifying assumption that cognate words always share
their meaning,\footnote{This is evidently false when considering the entire lexicon. There
  is a plethora of examples, such as as English \emph{deer} vs.\ German \emph{Tier}
  `animal', which are cognate (cf.\ \cite{kroonen12}, p.\ 94) without being
  synonyms. However, within the 40-concept core vocabulary space covered by ASJP, such
  cross-concept cognate pairs are arguably very rare.} the distribution of such similarity
scores for $c_1\neq c_2$ constitutes a sample of the overall distribution of similarity
scores between non-cognates.

Now consider the null hypothesis that the words for concept $c$ are non-cognate. We assume
\emph{a priori} that cognate word pairs are more similar than non-cognate ones. Let the
similarity score for $c$ be $x$. The maximum likelihood estimate for the $p$-value of that
null hypothesis is is the relative frequency non-cognate pairs with a similarity score
$\geq x$. If $PMI(c_i,c_j)$ is the similarity score between concept $c_i$ and $c_j$, we have
\begin{equation}\label{calib}
  p_c = \frac{|\{(c,c)\}\cup\{(c_i,c_j)|c_i\neq c_j \& PMI(c_i,c_j)\geq \mathit{PMI}(c,c)|}{%
    |\{(c,c)\}\cup\{(c_i,c_j)|c_i\neq c_j\}|}.
\end{equation}

Analogously to Fisher's method \cite{fisher1925}, the $p$-values for all concepts are combined
according to the formula
\begin{equation}
  \sum_c -\log p_c.
\end{equation}

If the null hypothesis is true for concept $c$, $p_c$ is distributed approximately according to a
continuous uniform distribution over the interval $(0,1]$. Accordingly, $-\log p_c$ is
distributed according to an exponential distribution with mean and variance
$=1$.\footnote{This follows from a standard argument about changes of variables in
  probability density functions. Let $f(x)$ be the density function for $y=-\log x$, where
  $x$ is distributed according to a uniform distribution over $(0,1]$. We have
  \begin{eqnarray*}
    \int_0^z f(y)dy & = & P(y\leq z)          \\
                    & = & P(-\log x\leq z)    \\
                    & = & P(\log x\geq -z)    \\
                    & = & 1-P(\log x < -z)    \\
                    & = & 1-P(x<e^{-z})       \\
                    & = & 1-e^{-z}            \\
    f(z)            & = & \frac{d(1-e^{-z})}{dz} \\
                    & = & e^{-z} 
  \end{eqnarray*}
} 

Suppose there are $N$ concepts for which both $l_1$ and $l_2$ have an entry. The sum of
$N$ independently distributed random variables, each with mean and variance $=1$,
approximately follows a normal distribution with mean $=N$ and variance $=N$. This can be
transformed into a $Z$-statistic by normalizing according to the formula
\begin{equation}\label{lsim}
  Z(l_1,l_2) = \frac{\sum_{i=1}^N -\log p_{c_i}-N}{\sqrt{N}}
\end{equation}

This normalization step addresses the third issue mentioned above, i.e., the varying
length of word lists.

$Z(l_1,l_2)$ increases with the degree of similarity between $l_1$ and $l_2$. It is
transformed into a dissimilarity measure\footnote{We will talk of \emph{distance measure}
  in the sequel for simplicity, even though it is not a metric distance.} as follows:
\begin{equation}\label{ldist}
  d(l_1,l_2) = \frac{Z_{\max}-Z(l_1,l_2)}{Z_{\max}-Z_{\min}}
\end{equation}

The maximal possible value $Z_{\max}$ for $Z$ would be achieved if both word lists have
the maximal length of $N=40$, and each synonymous word pair has a higher PMI score than
any non-synonymous word pair. Therefore
\begin{eqnarray*}
  Z_{\max} & =       & \frac{40\times -\log \frac{1}{40^2-40+1} -40}{\sqrt{40}} \\
          & \approx & 40.18
\end{eqnarray*}
The minimal value $Z_{\min}$ for $Z$ would be achieved if all $p_c$ equal 1 and both word
lists have length 40:
\begin{eqnarray*}
  Z_{\min} &=& \frac{40\times -\log 1 -40}{\sqrt{40}}\\
  &=& -\sqrt{40}\approx - 6.32
\end{eqnarray*}

We computed $d(l_1,l_2)$ for each pair of the above-mentioned 6,892 languages from the
ASJP database. This distance matrix is available \url|https://osf.io/24be8/|.

\subsection*{Automatic cognate classification}
\label{sec:cc}

\subsubsection*{Background}

In \cite{jaegerSofroniev16Konvens} a method is developed to cluster words into equivalence
classes in a way that approximates manual expert classifications. In this section this
approach is briefly sketched.

The authors chose a supervised learning approach. They use word lists with manual expert
cognate annotations from a diverse collection of language families, taken from
\cite{abvd,wichmannHolman13,list14Data,mennecieretal16,Dunn2012}. A part of these
goldstandard data were used to train a \emph{Support Vector Machine} (SVM). For each pair
of words $(w_1,w_2)$ from languages $(l_1,l_2$), denoting concept $c$, seven feature
values were computed:
\begin{enumerate}
\item \textbf{PMI similarity.} This is the string similarity measure according to
  \cite{jaeger13ldc} as described in the previous section.
\item \textbf{Calibrated PMI distance.} $p_c$ as defined in equation (\ref{calib}) above.
\item The negative logarithm thereof.
\item \textbf{Language similarity.} $Z(l_1,l_2)$, as defined in equation (\ref{lsim}) above.
\item The logarithm thereof.
\item \textbf{Average word length} of words for concept $c$ across all languages from the
  database, measured in number of symbols in ASJP transcription.
\item \textbf{Concept-language correlation.} The Pearson correlation coefficient between
  feature 3 and feature 4 for all word pairs expressing concept $c$.
\end{enumerate}

For each such word pair, the goldstandard contains an evaluation as \emph{cognate} (1) or
\emph{not cognate} (0). An SVM was trained to predict these binary cognacy
labels. Applying Platt scaling \cite{Platt1999}, the algorithm predics a \emph{probability
  of cognacy} for each pair of words from different languages denoting the same concept.
These probabilities were used as input for hierarchical clustering, yielding a
partitioning of words into equivalence classes for each concept.

The authors divided the goldstandard data into a training set and a test set. Using an SVM
trained with the training set, they achieve B-cubed F-scores \cite{bagga1998entity}
between $66.9\%$ and $90.9\%$ on the data sets in their test data, with a weighted average
of $71.8\%$ when comparing automatically inferred clusters with manual cognate
classifications.

\subsubsection*{Creating a goldstandard}

We adapted this approach to the task of performing automatic cognate classification on the
ASJP data. Since ASJP contains data from different families and it is confined to 40 core
concepts (while the data used in \cite{jaegerSofroniev16Konvens} partially cover 200-item
concept lists), the method has to be modified accordingly.

We created a goldstandard dataset from the data used in \cite{jaegerListSofroniev17}
(which is is drawn from the same sources as the data used in
\cite{jaegerSofroniev16Konvens} but has been manually edited to correct annotation
mistakes). Only the 40 ASJP concepts were used. Also, we selected the source data in such
a way that each dataset is drawn from a different language family. Words from different
families were generally classified as non-cognate in the goldstandard. All transcriptions
were converted into ASJP format. Table \ref{tab:1} summarizes the composition of the
goldstandard data.

\begin{table}[ht]
    \centering
    \resizebox{\linewidth}{!}{%
    \tabular{p{3cm}rrlp{3cm}lr}\toprule
    \bf Dataset & \bf Source           & \bf Words & \bf Concepts & \bf Languages & \bf Families     & \bf Cognate classes \\\midrule
    ABVD        & \cite{abvd}          & 2,306     & 34           & 100           & Austronesian     & 409                 \\
    Afrasian    & \cite{Militarev2000} & 770       & 39           & 21            & Afro-Asiatic     & 351                 \\
    Chinese     & \cite{Cihui}         & 422       & 20           & 18            & Sino-Tibetan     & 126                 \\
    Huon        & \cite{McElhanon1967} & 441       & 32           & 14            & Trans-New Guinea & 183                 \\
    IELex       & \cite{Dunn2012}      & 2,089     & 40           & 52            & Indo-European    & 318                 \\
    Japanese    & \cite{Hattori1973}   & 387       & 39           & 10            & Japonic          & 74                  \\
    Kadai       & \cite{Peiros1998}    & 399       & 40           & 12            & Tai-Kadai        & 102                 \\
    Kamasau     & \cite{Sanders1980}   & 270       & 36           & 8             & Torricelli       & 59                  \\
    Mayan       & \cite{brownetal08}   & 1,113     & 40           & 30            & Mayan            & 241                 \\
    Miao-Yao    & \cite{Peiros1998}    & 206       & 36           & 6             & Hmong-Mien       & 69                  \\
    Mixe-Zoque  & \cite{Cysouw2006}    & 355       & 39           & 10            & Mixe-Zoque       & 79                  \\
    Mon-Khmer   & \cite{Peiros1998}    & 579       & 40           & 16            & Austroasiatic    & 232                 \\
    ObUgrian    & \cite{Zhivlov2011}   & 769       & 39           & 21            & Uralic           & 68                  \\\midrule{}
    total       &                      & 10,106    & 40           & 318           & 13               & 2,311               \\\bottomrule
    \endtabular}
  \caption{Goldstandard data used for this study.}
\label{tab:1}
\end{table}

\subsubsection*{Clustering}

We used the \emph{Label Propagation} algorithm \cite{labelPropagation} for clustering. For
each concept, a network is constructed from the words for that concept. Two nodes are
connected if and only if their predicted probability of cognacy is $\geq
0.25$. \emph{Label Propagation} detects community structures within the network, i.e., it
partitions the nodes into clusters.

\subsubsection*{Model selection}

To identify the set of features suitable for clustering the ASJP data, we performed
\emph{cross-validation} on the goldstandard data. The data were split into a
\emph{training set}, consisting of the data from six randomly chosen language families,
and a \emph{test set}, consisting of the remaining data.  We slightly deviated from
\cite{jaegerSofroniev16Konvens} by replacing features 4 and 5 by \emph{language distance}
$d(l_1,l_2)$ as defined in equation (\ref{ldist}), and $-\log (1-d(l_1,l_2))$. Both are
linear transformations of the original features and therefore do not affect the automatic
classification.

For each of the 127 non-empty subsets of the seven features, an SVM with an RBF-kernel was
trained with 7,000 randomly chosen synonymous word pairs from the training
set.\footnote{Explorative tests revealed that accuracy of prediction does not increase if
  more training data are being used.} The trained SVM plus Platt scaling were used to
predict the probability of cognacy for each synonymous word pair from the test set, and
the resulting probabilities were used for Label Propagation clustering. This procedure was
repeated ten times for random splits of the goldstandard data into a training set and a
test set.

For each feature combination, the B-cubed F-score, averaged over the ten training/test
splits, was determined. The best performance (average B-cubed F-score: $0.86$) was
achieved using just two features:
\begin{itemize}
\item \textbf{Word similarity.} The negative logarithm of the calibrated PMI distance, and
\item \textbf{Language log-distance.} $-\log (1-d(l_1,l_2))$, with $d(l_1,l_2)$ as defined
  in equation (\ref{ldist}).
\end{itemize}

\begin{figure}[t]
  \centering
  \includegraphics[width=\linewidth]{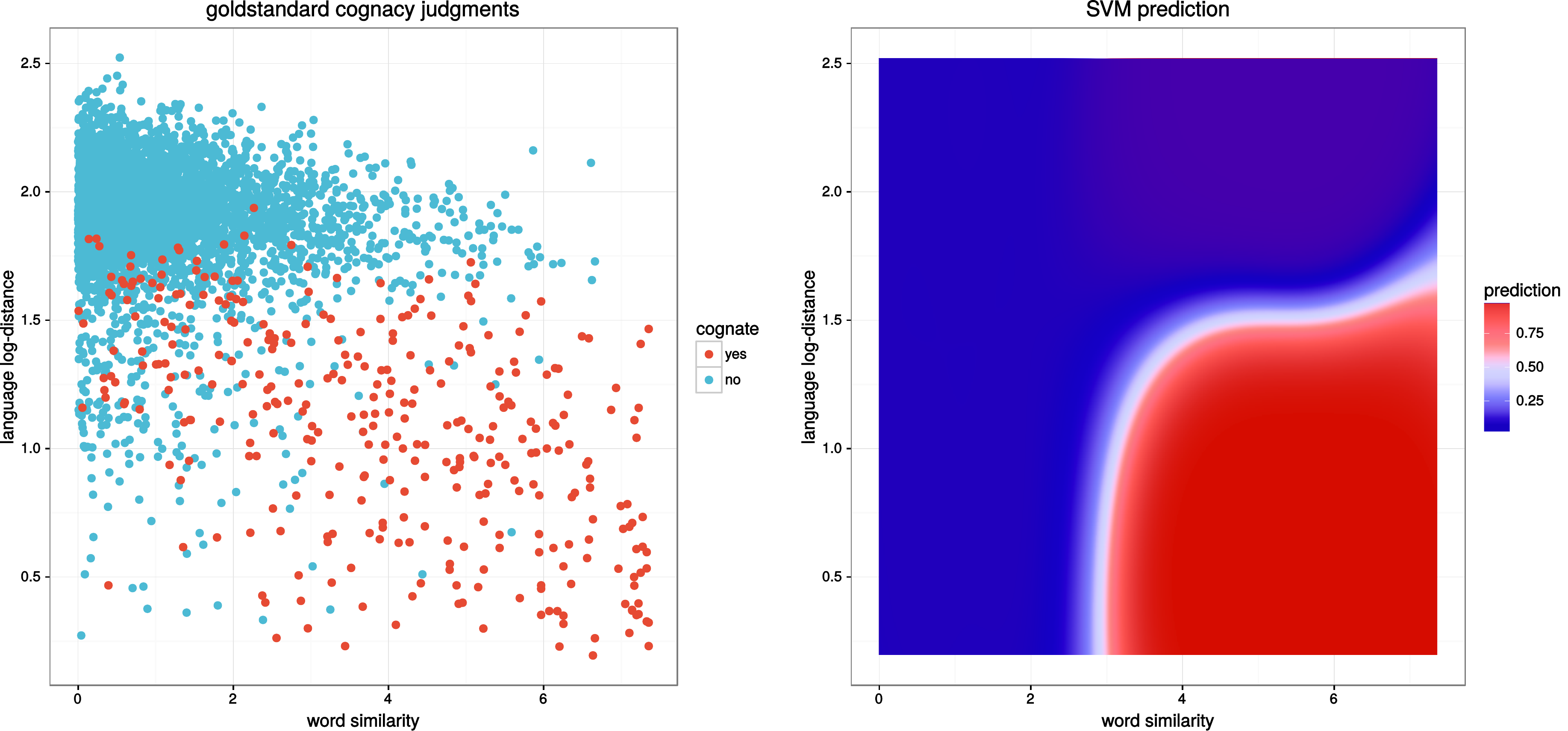}
  \caption{Expert cognacy judgments (left) and prediction of cognacy (right) depending on
    the selected features}
  \label{fig:2}
\end{figure}

Figure \ref{fig:2} displays, for a sample of goldstandard data, how expert cognacy
judgments depend on these features and how the trained SVM+Platt scaling predicts cognacy
depending on those features. Most cognate pairs are concentrated in the lower right corner
of the feature space, i.e., they display both high word similarity and low language
log-distance. The SVM learns this non-linear dependency between the two features.

\subsubsection*{Clustering the ASJP data}

A randomly selected sample of 7,000 synonymous word pairs from the goldstandard data were
used to train an SVM with an RBF-kernel, using the two features obtained via model
selection. Probabilities of cognacy for all pairs of synonymous pairs of ASJP entries were
obtained by (a) computing word similarity and language log-distance, (b) predict their
probability of cognacy using the trained SVM and Platt scaling, and (b) apply Label
Propagation clustering.

\subsection*{Phylogenetic inference}

\subsubsection*{Distance-based}

The language distances according to the definition in equation (\ref{ldist}) can be used
as input for distance-based phylogenetic infernce. In the experiments reported below, we
used the BIONJ \cite{bionj} algorithm for that purpose.

\subsubsection*{Character-based}

We propose two methods to extract discrete character matrices from the ASJP data.
\begin{enumerate}
\item \textbf{Automatically inferred cognate classes.} We defined one character per
  automatically inferred (in the sense described above) cognate class $cc$. If
  the word list for a language $l$ has a missing entry for the concept the elements of
  $cc$ refer to, the character is undefined for this language. Otherwise $l$ assumes value
  1 if its word list contains an element of $cc$, and 0 otherwise.
\item \textbf{Soundclass-concept characters.} We define a character for each pair $(c,s)$,
  where $c$ is a concept $c$ and $s$ an ASJP sound class. The character for $(c,s)$ is
  undefined for language $l$ if $l$'s word list has a missing entry for concept
  $c$. Otherwise $l$'s value is 1 if one of the words for $c$ in $l$ contains symbol $s$
  in its transcription, and 0 otherwise.
\end{enumerate}

The motivation for these two types of characters is that they track two different aspects
of language change. Cognacy characters contain information about lexical changes, while
soundclass-concept characters also track sound changes within cognate words. Both
dimensions provide information about language diversification.

Let us illustrate this with two examples.
\begin{itemize}
\item The Old English word for `dog' was \emph{hund}, i.e., \texttt{hund} in ASJP
  transcription. It belongs to the automatically inferred cognate class \emph{dog_149}. The
  Modern English word for that concept is \emph{dog}/\texttt{dag}, which belongs to class
  \emph{dog_150}. This amounts to two mutations of cognate-class characters between Old
  English and Modern English, $0\rightarrow 1$ for \emph{dog_150} and $1\rightarrow 0$ for
  \emph{dog_149}.

  The same historic process is also tracked by the sound-concept characters; it
  corresponds to seven mutations: $0\rightarrow 1$ for \emph{dog:d}, \emph{dog:a} and
  \emph{dog:g}, and $1\rightarrow 0$ for \emph{dog:h}, \emph{dog:u}, \emph{dog:n} and
  \emph{dog:d}.
\item The word for `tree' changed from Old English \emph{treow} (\texttt{treow}) to Modern
  English \emph{tree} (\texttt{tri}). Both entries belong to cognate class
  \emph{tree_17}. As no lexical replacement took place for this concept, there is no
  mutation of cognate-class characters separating Old and Modern English here. The
  historical sound change processes that are reflected in these words are captured by
  mutations of sound-concept characters: $0\rightarrow 1$ for \emph{tree:i} and
  $1\rightarrow 0$ for \emph{tree:e}, \emph{tree:o} and \emph{tree:w}.
\end{itemize}

For a given sample of languages, we use all \emph{variable} characters (i.e., characters
that have value 1 and value 0 for at least one language in the sample) from both sets of
characters. Phylogenetic inference was performed as Maximum-Likelihood estimation assuming
$\Gamma$-distributed rates with 25 rate categories, and using ascertainment bias
correction according to \cite{lewis2001}. Base frequencies and variance of rate variation
were estimated from the data.

In our phylogenetic experiments, the distance-based tree was used as initial tree for tree
search. This method was applied to three character matrices:
\begin{itemize}
\item cognate class characters,
\item soundclass-concept characters, and
\item a partitioned analysis using both types of characters simultaneously.
\end{itemize}
Inference was performed using the software RAxML \cite{raxml8}.

Applying more advanced methods of character-based inference, such as Bayesian inference
\cite{mrbayes3,beast,bayesphylogenies} proved to be impractical due to hardware
limitations.

\subsection*{Code availability}

The code used to conduct this study is freely available at
\mbox{\url|https://osf.io/cufv7/|} (DOI 10.17605/OSF.IO/CUFV7). The workflow processes
the sub-directories in the following order: 1.\ \texttt{pmiPipeline}, 2.\
\texttt{cognateClustering}, and 3.\ \texttt{validation}. All further details, including
software and software versions used, are described in the \texttt{README} files in the
individual sub-directories and sub-sub-directories (\url|https://osf.io/fskue/|,
\url|https://osf.io/2ncxm/|, \url|https://osf.io/32xg4/|, \url|https://osf.io/b8xgt/|, and
\url|https://osf.io/3rzfn/|).


\section*{Data Records}

All data that were produced are available at \url|https://osf.io/cufv7/| as well.

\subsection*{PMI data}

\begin{itemize}
\item estimated PMI scores and gap penalties: \mbox{\url|https://osf.io/rb3n4/|},\\
  \mbox{\url|https://osf.io/9bvfe/|} (csv files)
\item pairwise distances between languages: \url|https://osf.io/24be8/| (csv file)
\end{itemize}

\subsection*{Automatic cognate classification}

\begin{itemize}
\item word list with automatically inferred cognate class labels:\\
  \mbox{\url|https://osf.io/w52fu/|} (csv file)
\end{itemize}

\subsection*{Phylogenetic inference}

\begin{itemize}
\item family-wise data and trees (described in Subsection \emph{Phylogenetic Inference}
  within the Section \emph{Technical Validation}) are in sub-directory
  \texttt{validation/families} of \url|https://osf.io/cufv7/|. For each Glottolog family
  \texttt{F}, there are the following files (replace \texttt{[F]} by name of the family):
  \begin{itemize}
  \item {}\texttt{[F].cc.phy}: character matrix, cognate class characters, Phylip format
  \item {}\texttt{[F].sc.phy}: character matrix, soundclass-concept characters, Phylip
    format
  \item {}\texttt{[F].cc_sc.phy}: combined character matrix, cognate class and
    soundclass-concept characters, Phylip format
  \item {}\texttt{[F].part.txt}: partition file
  \item {}\texttt{[F].pmi.nex}: pairwise PMI distances, Nexus format
  \item \texttt{[F].pmi.tre}: BIONJ tree, inferred from PMI distances, Newick format
  \item \texttt{glot.[F].tre}: Glottolog tree, Newick format
  \item \texttt{RAxML_bestTree.[F]_cc}: Maximum Likelihood tree, inferred from cognate
    class characters, Newick format
  \item \texttt{RAxML_bestTree.[F]_sc}: Maximum Likelihood tree, inferred from
    soundclass-concept characters, Newick format
  \item \texttt{RAxML_bestTree.[F]_cc_sc}: Maximum Likelihood tree, inferred from
    combined character matrix, Newick format    
  \end{itemize}
\item global data over all 6,892 languages in the database are in the sub-directory
  \texttt{validation}, and global trees in the sub-directory
  \texttt{validation/worldTree}:
  \begin{itemize}
  \item \texttt{validation/world_cc.phy}: character matrix, cognate class characters, Phylip format
  \item \texttt{validation/world_sc.phy}: character matrix, soundclass-concept characters,
    Phylip format
  \item \texttt{validation/world_sc_cc.phy}: combined character matrix, cognate class and
    soundclass-concept characters, Phylip format
  \item \texttt{validation/world.partition.txt}: partition file
  \item \texttt{validation/glottologTree.tre}: Glottolog tree, Newick format
  \item \texttt{validation/worldTree/distanceTree.tre}: BIONJ tree, inferred from PMI
    distances, Newick format
  \item \texttt{validation/worldTree/RAxML_bestTree.world_cc}: Maximum Likelihood tree,
    inferred from cognate class characters, Newick format
  \item \texttt{validation/worldTree/RAxML_bestTree.world_sc}: Maximum Likelihood tree,
    inferred from soundclass-concept characters, Newick format
  \item \texttt{validation/worldTree/RAxML_bestTree.world_sc_cc}: Maximum Likelihood tree,
    inferred from combined character matrix, Newick format
  \item \texttt{validation/worldTree/RAxML_bestTree.world_sc_ccGlot}: Maximum Likelihood
    tree, inferred from combined character matrix using the Glottolog classifcation as
    constraint tree, Newick format
  \end{itemize}
\end{itemize}





\section*{Technical Validation}

\subsection*{Phylogenetic inference}

To evaluate the usefulness of the distance measure and the character matrices defined
above for phylogenetic inference, we performed two experiments:
\begin{itemize}
\item \textbf{Experiment 1.} We applied both distance-based inference and character-based
  inference for all language families (according to the Glottolog classification)
  containing at least 10 languages in ASJP.
\item \textbf{Experiment 2.} We sampled 100 sets of languages with a size between 20 and
  400 at random and applied all four methods of phylogenetic inference to each of them.
\end{itemize}

In both experiments, each automatically inferred phylogeny was evaluated by computing the
\emph{Generalized Quartet Distance} (GQD) \cite{pompeieatl11} to the Glottolog expert tree
(restricted to the same set of languages).

\begin{figure}
  \centering
  \includegraphics[width=1\linewidth]{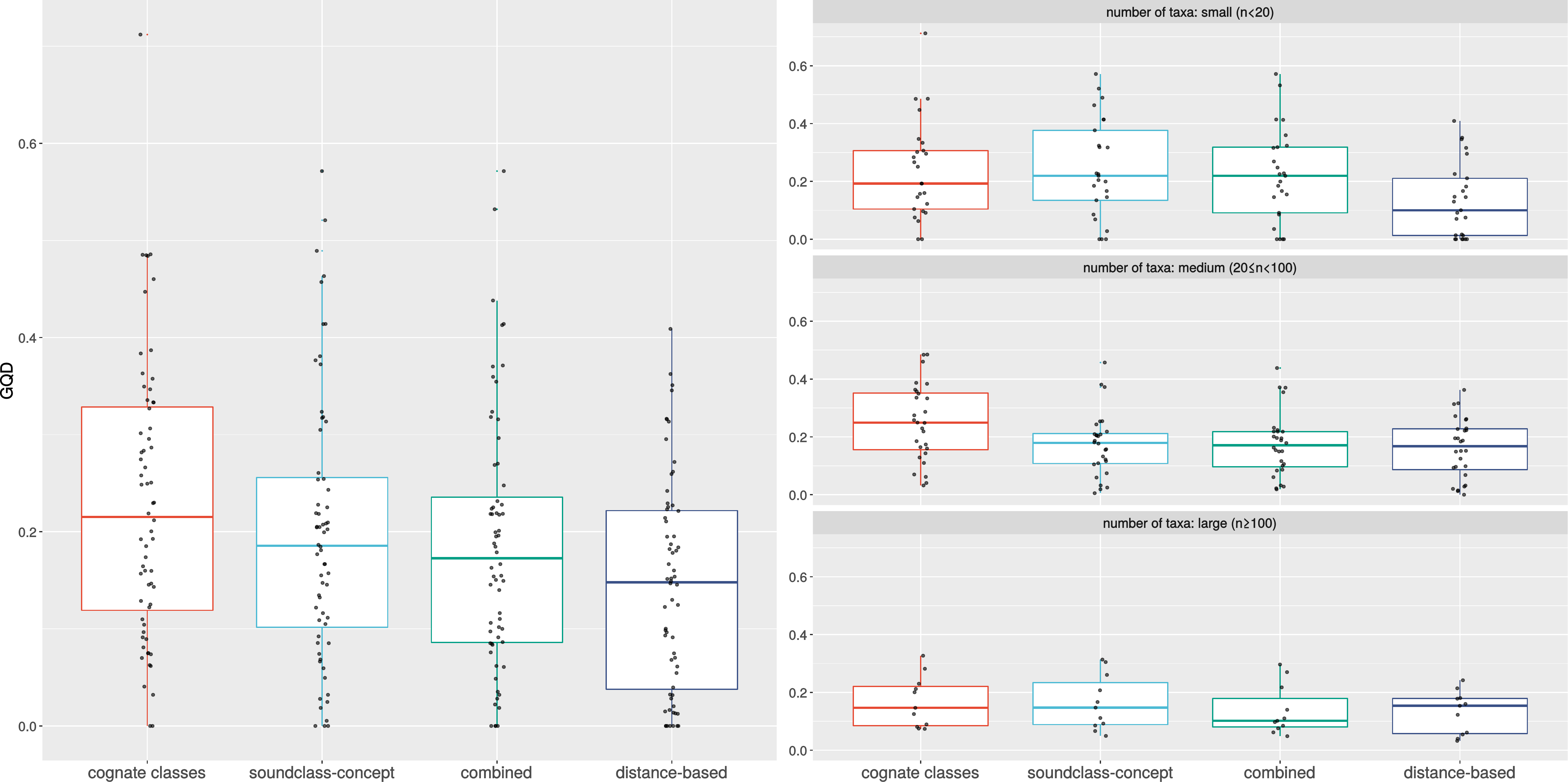}
  \caption{Experiment 1: Generalized Quartet Distances for Glottolog families depending on
    phylogenetic inference method. Aggregated over all families (left) and split according
  to family size (right).}
  \label{fig:3}
\end{figure}

The results of the first experiment are summarized in Table \ref{tab:2} and visualized in
Figure \ref{fig:3}. The results for the individual families are given in Table \ref{tab:a1}.

\begin{table}[t]
  \centering
  \resizebox{\linewidth}{!}{%
    \begin{tabular}{lcccc}
      \toprule
      \emph{method}                     & \multicolumn{3}{c}{character-based} & distance-based                        \\
      \emph{character type}             & cognate classes                     & soundclass-concept & combined &       \\\midrule
      total                             & 0.215                               & 0.186              & 0.173    & 0.148 \\\midrule[.5pt]
      small families ($n<20$)           & 0.193                               & 0.219              & 0.219    & 0.100 \\
      medium families ($20\leq n< 100$) & 0.249                               & 0.179              & 0.171    & 0.168 \\
      large families ($n\geq 100$)      & 0.147                               & 0.147              & 0.102    & 0.154 \\\bottomrule
    \end{tabular}
  }
  \caption{Median Generalized Quartet Distances for Glottolog families}
  \label{tab:2}
\end{table}
Aggregating over all families suggests that distance-based inference produces the best fit
with the expert goldstandard. However, a closer inspection of the results reveals that the
performance of the different phylogenetic inference methods depend on the size of the
language families (measured in number of taxa available in ASJP). Combining both types of
characters in a partitioned model always leads to better results than the two character
types individually. While distance-based inference is superior for small language families
(less than 20 taxa), character-based inference appears to be about equally good for
medium-sized (20--199 taxa) and large (more than 200 taxa) language families.

\begin{figure}
  \centering
  \includegraphics[width=1\linewidth]{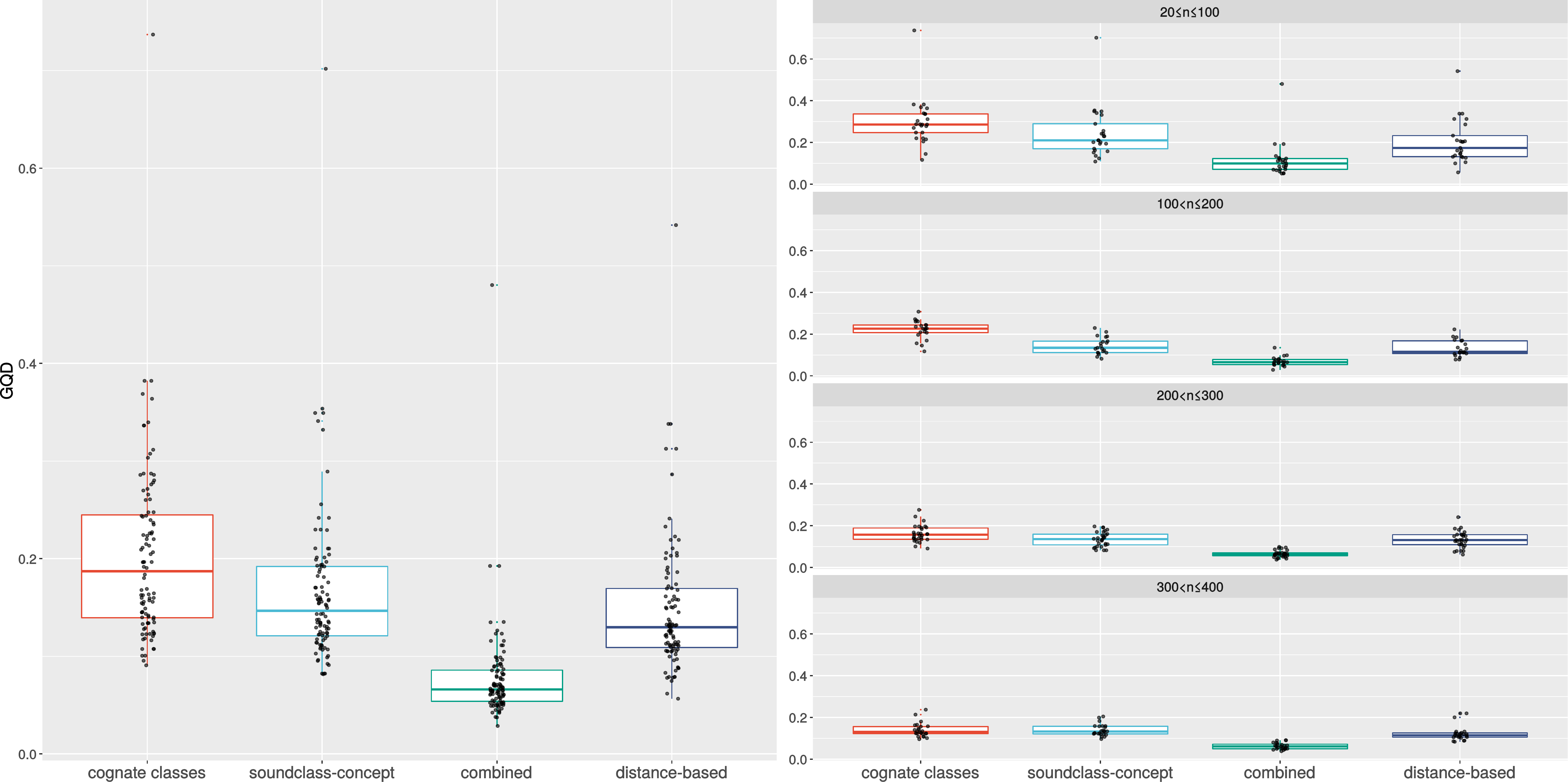}
  \caption{Experiment 2: Generalized Quartet Distances for random samples of languages
    depending on phylogenetic inference method. Aggregated over all samples (left) and
    split according to sample size (right).}
  \label{fig:4}
\end{figure}

This assessment is based on a small sample size since there are only 33 medium-sized and
6 large language families. The results of experiment 2 confirm these conclusions
though. They are summarized in Table \ref{tab:3} and illustrated in Figure \ref{fig:4}.
\begin{table}[t]
  \centering
  \resizebox{\linewidth}{!}{%
    \begin{tabular}{lcccc}
      \toprule
      \emph{method}         & \multicolumn{3}{c}{character-based} & distance-based                        \\
      \emph{character type} & cognate classes                     & soundclass-concept & combined &       \\\midrule
      total                 & 0.187                               & 0.147              & 0.066    & 0.130 \\\midrule[.5pt]
      $~20\leq n\leq 100$   & 0.286                               & 0.210              & 0.099    & 0.174 \\
      $100< n\leq 200$      & 0.226                               & 0.135              & 0.065    & 0.115 \\
      $200<n\leq 300$       & 0.157                               & 0.136              & 0.063    & 0.132 \\
      $300<n\leq 400$       & 0.131                               & 0.132              & 0.061    & 0.114 \\\bottomrule
    \end{tabular}
  }
  \caption{Median Generalized Quartet Distances to Glottolog for random samples of
    languages}
  \label{tab:3}
\end{table}
All four methods improve with growing sample size, but this effect is more pronounced for
character-based inference. While combined character-based inference and distance-based
inference are comparable in performance for smaller samples of languages ($n\leq 100$),
character-based inference outperforms distance-based inference for larger samples, and the
difference grows with sample size.

The same pattern is found when the different versions of phylogenetic inference is applied
to the full dataset of 6,892 languages. We find the following GQD values:
\begin{itemize}
\item distance based tree (\url|https://osf.io/vy456/|): $0.078$
\item cognate-class based ML tree (\url|https://osf.io/d89cr/|): $0.052$
\item soundclass-concept based ML tree (\url|https://osf.io/uh6rf/|): $0.089$
\item ML tree from combined character data (\url|https://osf.io/dg5jh/|): $0.035$
\end{itemize}

\subsection*{Relation to geography}

Both the distances between languages and the two methods to represent languages as
character vectors are designed to identify similarities between word lists. There are
essentially three conceivable causal reasons why the word lists from two languages are
similar: (1) common descent, (2) language contact and (3) universal tendencies in
sound-meaning association due to sound symbolism, nursery forms etc.\
\cite{blasietal2016}. The third effect is arguably rather weak. The signal derived from
common descent and from language contact should be correlated with geographic distance. If
the methods proposed here extract a genuine signal from word lists, we thus expect to find
such a correlation.

To test this hypothesis, we computed the geographic distance (great-circle distance)
between all pairs from a sample of 500 randomly selected languages, using the geographic
coordinates supplied with the ASJP data.

We furthermore extracted pairwise distances from character vectors by computing the cosine
distance between those vectors, using only characters for which both languages have a
defined value. In this way we obtained three matrices of pairwise linguistic distances for
the mentioned sample of 500 languages: (1) The distance as defined in equation
(\ref{ldist}), called \textbf{PMI distance}, (2) the cosine distance between the
cognate-class vectors, and (3) the cosine distance between the sound-concept vectors.

\begin{figure}[ht]
  \centering
  \includegraphics[width=\linewidth]{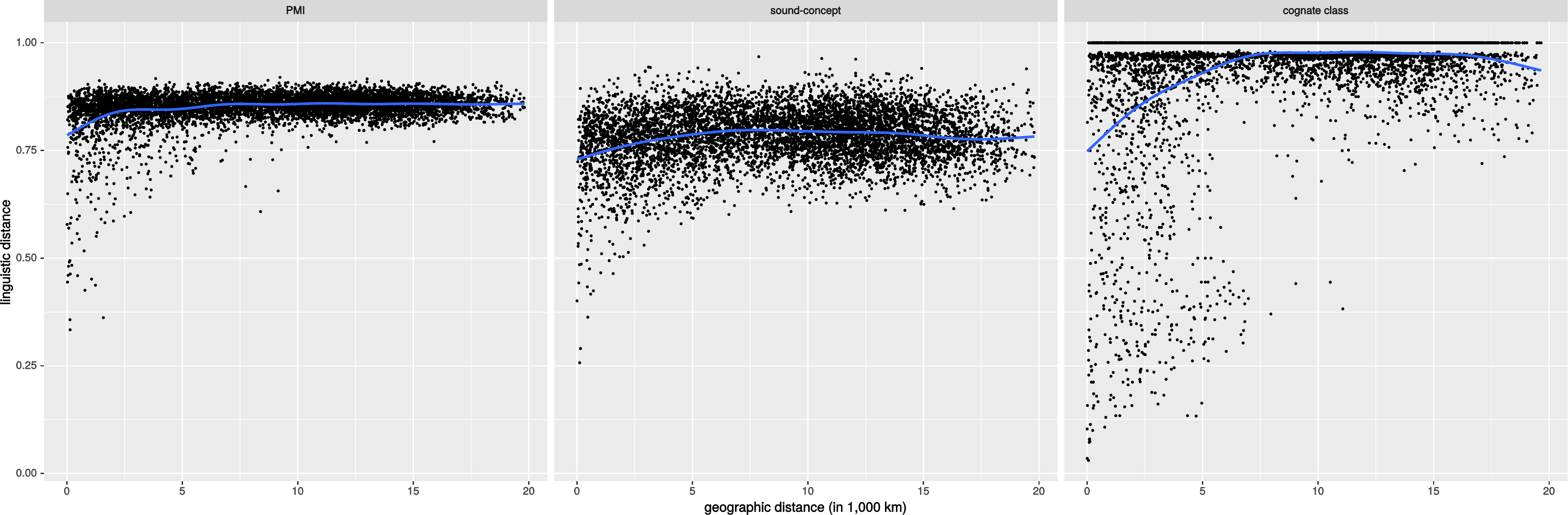}
  \caption{Geographic vs.\ linguistic distances between languages}
  \label{fig:5}
\end{figure}

All three linguistic distance measures show a signficant correlation with geographic
distance. The Pearson correlation coefficient for PMI distances is $0.252$ (p=$0.001$
according to the Mantel test), $0.329$ (p=$0.001$) for cognate-class distance and $0.140$
($p=0.001$) for sound-concept distance. Figure \ref{fig:5} shows the corresponding scatter
plots.

\begin{figure}[ht]
  \centering
  \includegraphics[width=\linewidth]{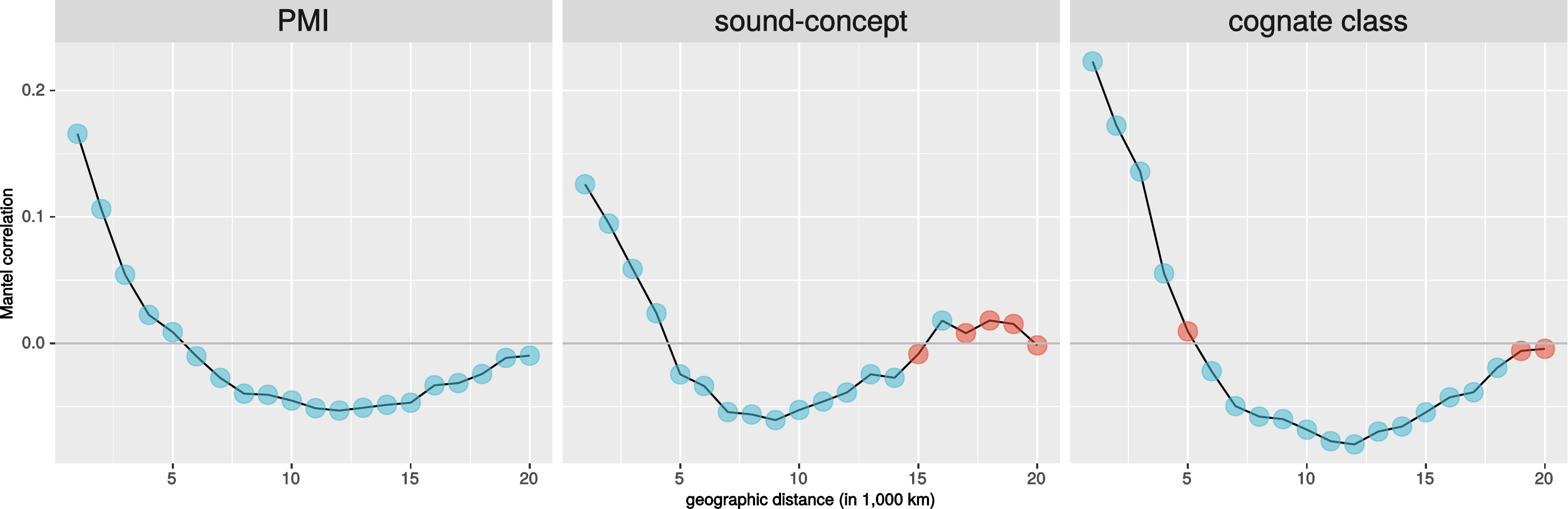}
  \caption{Mantel correlograms. Blue: significant, red: non-significant at $p<0.05$.}
  \label{fig:6}
\end{figure}

The visualization suggests that for all three linguistic distance measures, we find a
signal at least up to 5,000 km. This is confirmed by the Mantel correlograms
\cite{legendreLegendre2012} shown in Figure \ref{fig:6}.
We find a significant positive correlation with geographic distance for up to 5,000 km for
PMI distance, and up to 4,000 km for cognate-class distance and sound-concept distance.

\section*{Usage Notes}

Character-based inference from expert cognacy judgment data have been used in various
downstream applications beyond phylogenetic inference, such as estimating the time course
of prehistoric population events
\cite{grayJordan2000,grayDrummondGreenhill09,bouckaertetal12} or the identification of
overarching patterns of cultural language evolution
\cite{pageletal2007,atkinsonetal2008}. In this section it will be illustrated how the
automatically inferred characters described above can be deployed to expand the scope of
such investigations to larger collections of language families.

\subsection*{A case study: punctuated language evolution}

A few decades ago, \cite{gouldEldredge77} proposed that biological evolution is not, in
general, a gradual process. Rather, they propose, long periods of stasis are separated by
short periods of rapid change co-occurring with branching speciation. This model goes by
the name of \emph{punctuated equilibrium}. This proposal has initiated a lively and still
ongoing discussion in biology. Pagel, Venditti and Meade \cite{pageletal2006} developed a
method to test a version of this hypothesis statistically. They argue that most
evolutionary change occurs during speciation events. Accordingly, we expect a positive
correlation between the number of speciation events a lineage underwent throughout its
evolutionary history and the amount of evolutionary change that happened during that
time. 

Estimates of both quantities can be read off a phylogenetic tree --- the number of
speciation events corresponds to the number of branching nodes, and the amount of change
to the total path length --- provided (a) the tree is rooted and (b) branch lengths
reflect evolutionary change (e.g., the expected number of mutations of a character) rather
than historical time. In \cite{pageletal2006} a significant correlation is found for
biomolecular data, providing evidence for punctual evolution.

In \cite{atkinsonetal2008}, the same method is applied to the study of language evolution,
using expert cognacy data from three language families (Austronesian, Bantu,
Indo-European). The study results in strong evidence for punctuated evolution in all three
families. 

We conducted a similar study for all Glottolog language families with at least 10 ASJP
languages. The workflow was as follows. For each family $F$:
\begin{itemize}
\item Find the language $o\not\in F$ which has the minimal average PMI distance to the
  languages in $F$. This language will be used as \emph{outgroup}.
\item Infer a Maximum-Likelihood tree over the taxa $F\cup\{o\}$ with the Glottolog
  classification as constraint tree, using a partitioned analysis with cognate-class
  characters and soundclass-concept characters.
\item Use $o$ as outgroup to root the tree; remove $o$ from the tree.
\item Apply the \emph{$\delta$-test} \cite{vendittietal2006} to control for the \emph{node
    density artifact}. 
\item Perform \emph{Phylogenetic Generalized Least Square} \cite{grafen89} regression with
  root-to-tip path lengths for all taxa as independent and root-to-tip number of nodes as
  dependent variable.
\item If the $\delta$-test is negative and the regression results in a significantly
  positive slope, there is evidence for punctuated evolution in $F$.
\end{itemize}

Among the 66 language families considered, the $\delta$-test was negative for 43
families. We applied Holm-Bonferroni correction for multiple testing to determine
significance in the regression analysis. The numerical results are given in Table
\ref{tab:punctuation}. 
\begin{table}[t]
  \centering
  \begin{tabular}{lrrrl}
    \toprule
    family                   & slope  & $p$-value & number of taxa & significant \\\midrule
    Atlantic-Congo           & 0.003  & $<$1E-14  & 1332           & yes         \\
    Austronesian             & 0.005  & $<$1E-14  & 1259           & yes         \\
    Afro-Asiatic             & 0.008  & 2E-13     & 356            & yes         \\
    Sino-Tibetan             & 0.005  & 9E-8      & 279            & yes         \\
    Indo-European            & 0.004  & 2E-7      & 367            & yes         \\
    Nuclear_Trans_New_Guinea & 0.003  & 7E-4      & 259            & yes         \\
    Pama-Nyungan             & 0.005  & 6E-4      & 167            & yes         \\\midrule
    Tai-Kadai                & 0.007  & 8E-3      & 142            & no          \\
    Kiwaian                  & 0.024  & 9E-3      & 10             & no          \\
    Nakh-Daghestanian        & 0.011  & 0.01      & 55             & no          \\
    Turkic                   & 0.009  & 0.02      & 60             & no          \\
    Quechuan                 & 0.006  & 0.03      & 62             & no          \\
    Siouan                   & 0.004  & 0.04      & 17             & no          \\
    Cariban                  & 0.016  & 0.05      & 30             & no          \\
    Eskimo-Aleut             & -0.052 & 0.05      & 10             & no          \\
    Central_Sudanic          & 0.010  & 0.07      & 58             & no          \\
    Salishan                 & 0.015  & 0.08      & 30             & no          \\
    Chibchan                 & 0.011  & 0.10      & 23             & no          \\
    Ainu                     & 0.013  & 0.10      & 22             & no          \\
    Dravidian                & 0.008  & 0.10      & 38             & no          \\
    Sko                      & 0.038  & 0.10      & 14             & no          \\
    Uralic                   & 0.017  & 0.12      & 30             & no          \\
    Ndu                      & 0.025  & 0.14      & 10             & no          \\
    Lower_Sepik-Ramu         & 0.070  & 0.15      & 19             & no          \\
    Japonic                  & 0.010  & 0.17      & 32             & no          \\
    Gunwinyguan              & 0.027  & 0.21      & 14             & no          \\
    Heibanic                 & -0.041 & 0.24      & 11             & no          \\
    Khoe-Kwadi               & 0.015  & 0.39      & 12             & no          \\
    Tungusic                 & 0.011  & 0.40      & 25             & no          \\
    Tucanoan                 & -0.011 & 0.45      & 32             & no          \\
    Angan                    & 0.018  & 0.46      & 17             & no          \\
    Cochimi-Yuman            & 0.005  & 0.47      & 13             & no          \\
    Chocoan                  & -0.027 & 0.51      & 10             & no          \\
    Kadugli-Krongo           & -0.006 & 0.71      & 11             & no          \\
    Pano-Tacanan             & 0.001  & 0.78      & 33             & no          \\
    Tupian                   & 0.001  & 0.80      & 59             & no          \\
    Totonacan                & -0.002 & 0.80      & 14             & no          \\
    Ta-Ne-Omotic             & -0.003 & 0.83      & 24             & no          \\
    Algic                    & -0.009 & 0.86      & 32             & no          \\
    Lakes_Plain              & -0.002 & 0.89      & 22             & no          \\
    Timor-Alor-Pantar        & 0.005  & 0.92      & 59             & no          \\
    Bosavi                   & -0.003 & 0.99      & 13             & no          \\
    \bottomrule
  \end{tabular}
  \caption{Test for punctuated language evolution for the families without node density
    artifact. Significance is determined via Holm-Bonferroni correction at the
    significance level of $0.05$.}
  \label{tab:punctuation}
\end{table}

\begin{figure}[ht]
  \centering
  \includegraphics[width=\linewidth]{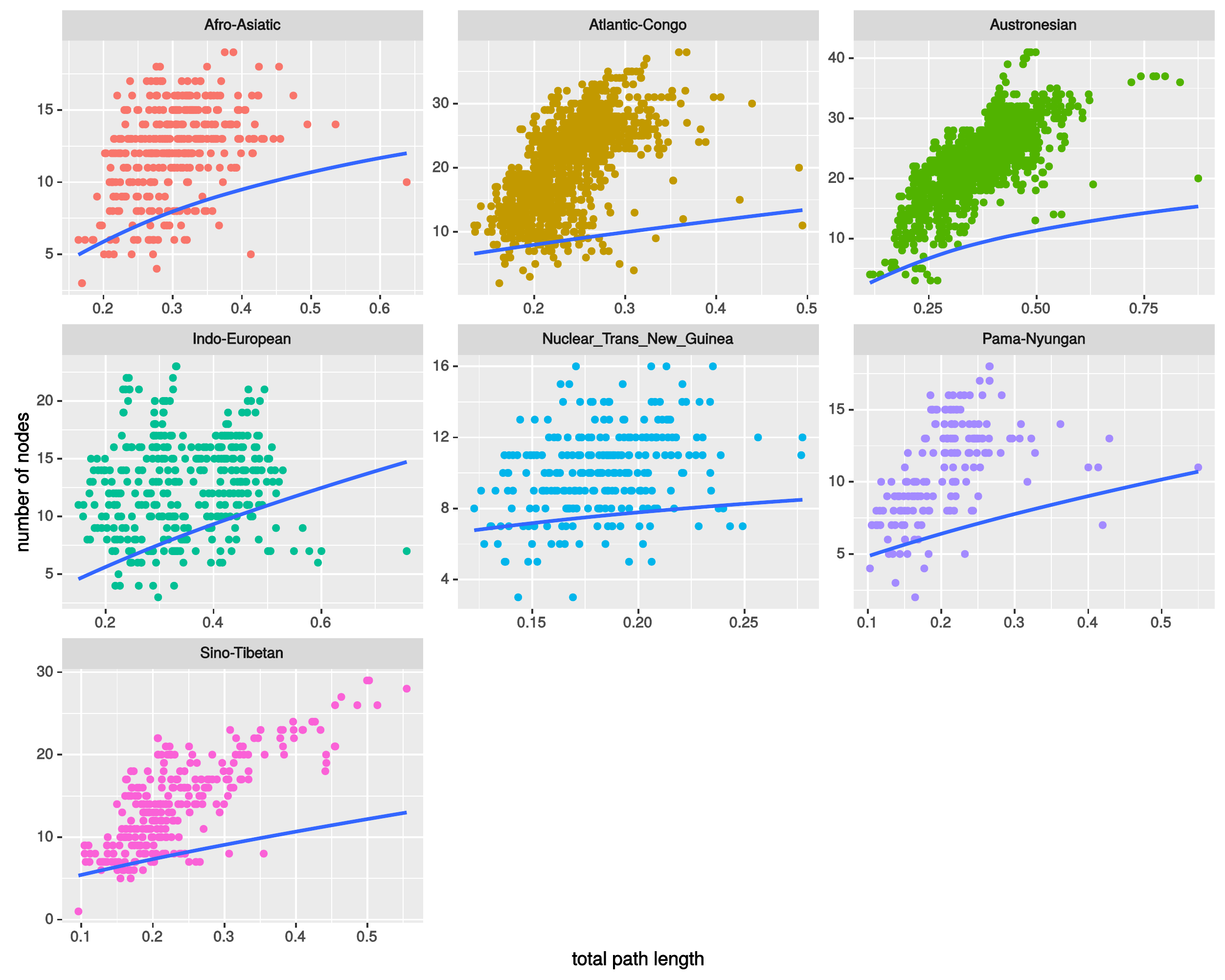}
  \caption{Dependency between total path length and the number of branching nodes for the
    families with a significantly positive association. Blue lines are regression lines
    according to phylogenetic generalized least squares, using $\delta$-correction.}
  \label{fig:7}
\end{figure}

A significant positive dependency was found for the seven largest language families
(Atlantic-Congo, Austronesian, Indo-European, Afro-Asiatic, Sino-Tibetan, Nuclear
Trans-New Guinea, Pama-Nyungan). The relationships for these families are visualized in
Figure \ref{fig:7}. No family showed a significant negative dependency. This strengthend
the conclusion of Atkinson et al.\ \cite{atkinsonetal2008} that languages evolve in
punctuational bursts.




\section*{Acknowledgements}

This research was supported by the ERC Advanced Grant 324246 EVOLAEMP and the DFG-KFG 2237
\emph{Words, Bones, Genes, Tools}, which is gratefully acknowledged.

\begin{table}[p]
  \centering\small
  \includegraphics[width=\linewidth]{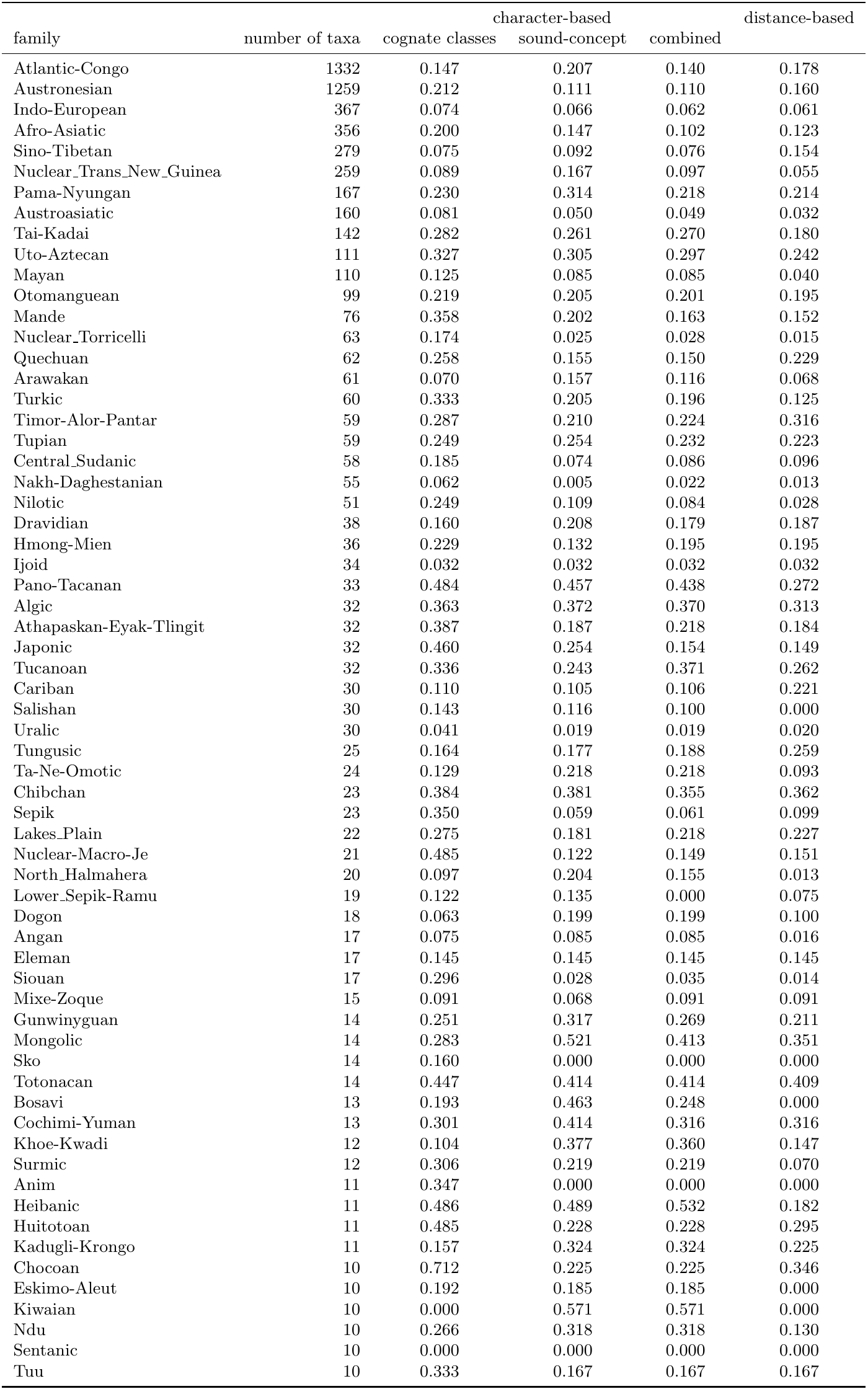}
  \caption{Generalized Quartet Distances to Glottolog expert tree for Glottolog families
    with $\geq$ 10 taxa}
  \label{tab:a1}
\end{table}

\begin{table}[p]
  \centering\small
  \includegraphics[width=\linewidth]{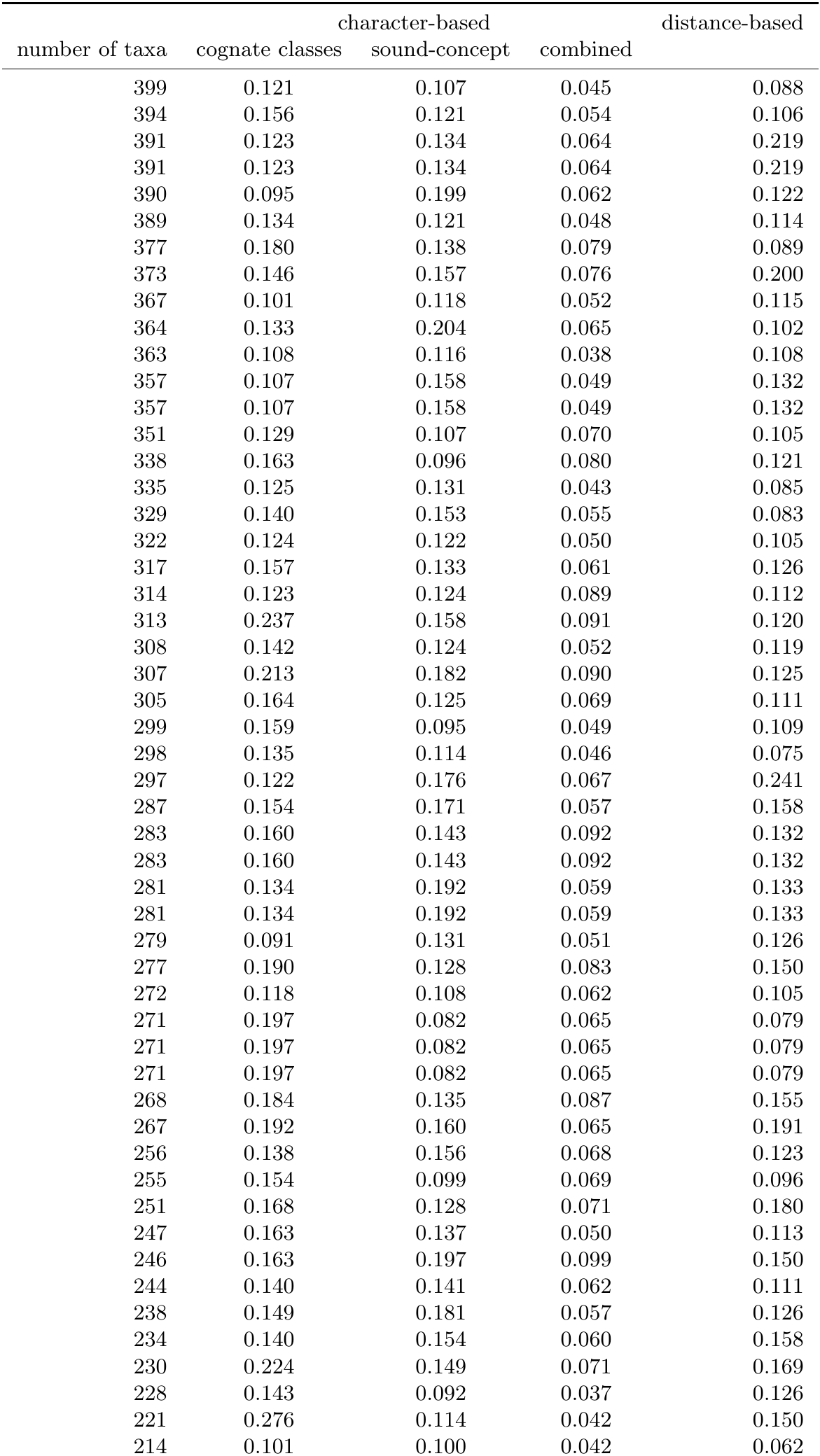}
\end{table}
\begin{table}[p]
  \includegraphics[width=\linewidth,page=2]{table2Cropped}
  \caption{Generalized Quartet Distances to Glottolog expert tree for randomly sampled
    groups of taxa}
  \label{tab:a2}
\end{table}








\end{document}